\documentclass[runningheads]{llncs}

\usepackage{accv}

\usepackage{accvabbrv}
\usepackage{graphicx}
\usepackage{amsmath}
\usepackage{mathtools}
\usepackage{booktabs}
\usepackage{multirow}
\usepackage{siunitx}
\usepackage{algorithm}
\usepackage{algpseudocode}
\usepackage{hyperref}
\hypersetup{hidelinks}
\usepackage[accsupp]{axessibility}

\begin{document}
\title{VLM-Aware Meta-Optic Front-End Design for Frozen Vision-Language Models}
\titlerunning{VLM-Aware Meta-Optic Front-End Design}

\author{Chanik Kang\inst{1, 2} \and
Raphaël Pestourie\inst{2,*}\and
Haejun Chung \inst{1,3,*}}

\authorrunning{Kang et al.}

\institute{Department of Artificial Intelligence, Hanyang University, Seoul, 04763, Korea
\and
School of Computational Science and Engineering, Georgia Institute of Technology, Atlanta, GA, 30332, USA\\
\email{rpestourie3@gatech.edu}
\and
Department of Electronic Engineering, Hanyang University, Seoul, 04763, Korea\\
\email{haejun@hanyang.ac.kr} \\
\textsuperscript{*}Corresponding authors
}

\maketitle

\begin{abstract}
Conventional machine-vision pipelines typically rely on high-quality optics that produce clean, human-interpretable images, and optical design has therefore been driven by image-level criteria such as resolution, aberration correction, and pixel fidelity. However, such optics are often impractical for size-, cost-, or form-factor-constrained applications, where compact meta-optics offer an attractive alternative but operate under strict physical efficiency limits. We propose CODA, a co-design framework that optimizes a continuous-density meta-optic front-end for frozen-model recognition using differentiable image formation and adjoint-gradient updates of Maxwell-based simulations. CODA directly optimizes the cross-entropy loss of a fixed zero-shot CLIP classifier without learned reconstruction, image signal processing, or image-fidelity auxiliary objectives. In a two-dimensional simulated imaging benchmark on ImageNet-100, CODA improves CLIP ViT-L/14 zero-shot accuracy from 53.75 $\pm$ 3.57\% with a focal-concentration baseline to 65.41 $\pm$ 3.99\%. The optimized optics further transfer without re-optimization across CLIP, SigLIP, and DINOv2 on ImageNet-100, CIFAR-100, and Food-101. These results demonstrate that, under constrained meta-optic imaging, downstream recognition can be improved by aligning optical design with frozen vision-model objectives rather than conventional image-formation criteria.

\end{abstract}

\section{Introduction}
\label{sec:intro}
Large-scale vision--language models (VLMs) are increasingly used as general-purpose visual consumers for open-vocabulary classification, retrieval, and other perception tasks~\cite{radford2021learning,jia2021scaling,zhai2023siglip}. When clean, conventionally rendered images are available, they are the natural input interface for these models: their visual encoders are primarily trained and evaluated on this image domain. We therefore do not frame optical co-design as a way to outperform a high-quality clean-image interface. Instead, we study the constrained case: when the optical front-end must be compact or multifunctional, can it be optimized to produce sensor measurements that are more useful to a frozen VLM?

\begin{figure}[!t]
 \centering
 \includegraphics[width=\linewidth]{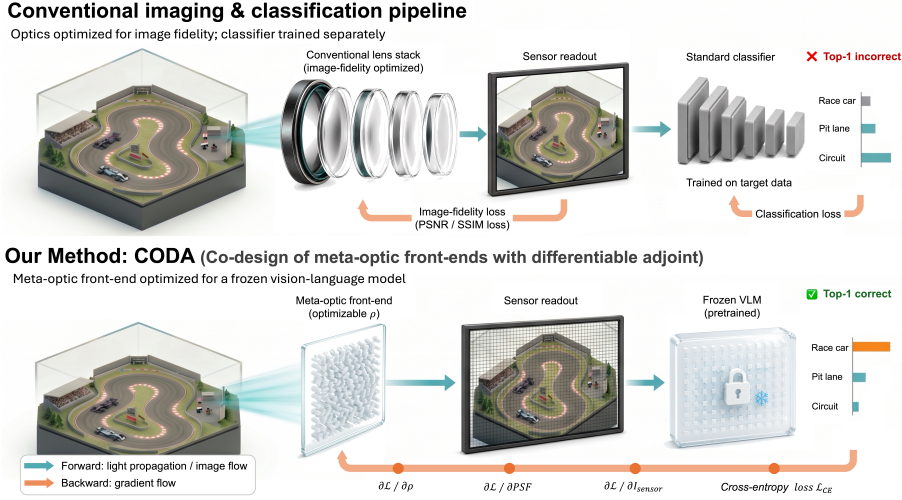}
\caption{
\textbf{Schematic overview of a conventional imaging-classification pipeline and CODA}
(co-design of meta-optic front-ends with differentiable adjoints).
CODA keeps the VLM fixed and optimizes only the meta-optic density $\rho$ for recognition. The classification loss is differentiated through the frozen VLM and image-formation model to the point-spread-function (PSF) interface, and an adjoint Maxwell solve yields $\partial\mathcal{L}/\partial\rho$.
}
 \label{fig:fig1}
\end{figure}

Conventional imaging is not always the most practical hardware interface. In applications where thickness, weight, aperture sharing, manufacturability, or multifunctional wavefront control dominates the system budget, meta-optic front-ends can offer substantial advantages. Their planar subwavelength structures can replace bulky multi-element lens stacks while enabling compact angle-, wavelength-, or polarization-dependent responses. The challenge is that these advantages often come with constrained image formation. In high-numerical-aperture~\cite{chung2020high,chen20253d}, wide-field-of-view~\cite{martins2020metalenses,liu2024ultra,wirth2025wide}, and ultra-compact meta-optic settings~\cite{fu2022ultracompact,seo2024metalens,wang2024quantitative,LAM2026030080,hao2026compact}, the sensor measurement may be shaped by aberration, diffraction, multiplexing, or other departures from clean image formation~\cite{liang2019high,colburn2018metasurface,tseng2021neural,shen2023monocular,zhang2024neural}. Such optical inputs are still often evaluated using human-interpretable criteria such as resolution, contrast, and reconstruction fidelity~\cite{zhou2011computational,tseng2021differentiable,sitzmann2018endtoend}.

This creates a mismatch for foundation-model inference. Image-fidelity metrics such as peak signal-to-noise ratio (PSNR) and structural similarity index measure (SSIM) reward visually faithful reconstructions~\cite{wang2004image}, while focusing objectives reward compact point-spread functions. A frozen VLM, however, maps its input into a learned representation space and may not rank constrained optical measurements by the same criteria. Thus, in a constrained optical regime, the front-end preferred by conventional image-formation metrics need not be the front-end preferred by downstream recognition.

Prior end-to-end computational imaging systems have shown that optical front-ends can be optimized for downstream computational objectives~\cite{chang2018hybrid,lin2022end,lin2021end,zhou2011computational,tseng2021neural,tseng2021differentiable, moclip2025}. In many such systems, however, the optics are optimized together with a reconstruction network, image-processing module, or task-specific neural backend. This makes the source of any gain ambiguous: it may come from the optical front-end, from a backend adapted to optical artifacts, or from both. Foundation-model deployment presents a different setting: the downstream model is already trained and is often used as a fixed visual consumer. Accordingly, we ask whether changing only the optical front-end can better serve a frozen VLM.

We propose CODA (co-design of meta-optic front-ends with differentiable adjoints), a framework for optimizing a constrained meta-optic front-end with a frozen VLM objective. Here, ``front-end'' refers to the meta-optic imaging element and its differentiable sensor-image formation model, not a complete camera system including the image sensor, mechanical package, or image signal processor. CODA optimizes only the continuous density of a planar meta-optic while keeping the VLM and class text embeddings fixed. For each wavelength--angle condition, a Maxwell-equation simulation produces a one-dimensional point-spread function (PSF), which is used to synthesize sensor images through a differentiable line-scan image-formation model. The frozen VLM consumes these sensor images, and its classification loss is differentiated back to the PSF interface and then converted into an adjoint-gradient update of the meta-optic density. No learned deconvolution, reconstruction network, or image signal processing module is inserted between the optic and the frozen model.

We make the following empirical and methodological contributions. First, we formulate frozen-VLM optical front-end optimization as a single-objective problem driven by downstream classification loss, without image-fidelity auxiliary losses. Second, we connect frozen-encoder automatic differentiation to adjoint-gradient updates of a Maxwell-based meta-optic simulation through a PSF backward interface. Third, within the same simulated meta-optic design domain and image-formation model, we show that switching from a model-agnostic focal-concentration objective to the frozen-VLM objective improves CLIP ImageNet-100 accuracy by $+11.66$ percentage points. Finally, we show that the same optimized optics, without optical re-optimization, outperform the focal-concentration baseline across CLIP~\cite{radford2021learning}, SigLIP~\cite{zhai2023siglip}, and DINOv2~\cite{oquab2024dinov2} evaluations on ImageNet-100~\cite{deng2009imagenet, tian2020crd}, CIFAR-100~\cite{krizhevsky2009learning}, and Food-101~\cite{bossard14}. We do not claim that CODA surpasses clean-image inputs or improves every optical design problem; rather, our results show that VLM-aware optical optimization can improve recognition when the front-end is constrained and the downstream visual model is frozen.

\section{Positioning relative to prior work}
\label{sec:positioning}

\begin{figure}[!htp]
 \centering
 \includegraphics[width=\linewidth]{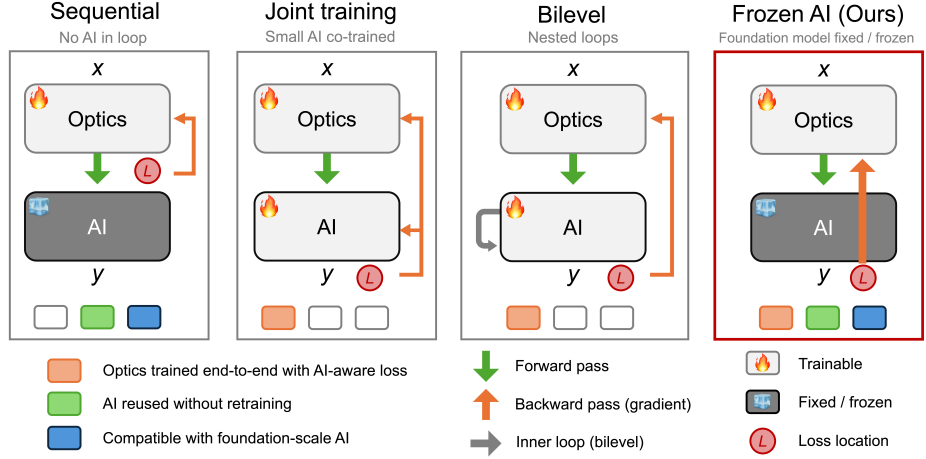}
\caption{
\textbf{Representative optics-adaptation formulations for optics--AI co-design.}
Green/orange arrows denote forward/backward passes, and light/dark blocks indicate trainable/frozen components.
Unlike sequential, joint, and bilevel formulations, CODA freezes the visual foundation model and back-propagates its classification loss only to the meta-optic density.
}
\label{fig:fig2}
\end{figure}

Our setting differs from three adjacent lines of work. First, meta-optics inverse design uses gradient-based electromagnetic inverse design, often implemented with adjoint simulations, to optimize large geometric or material degrees of freedom~\cite{ma2026inverse,molesky2018inverse,hughes2018adjoint,hammond2022meep,meem2021imaging}. These methods typically optimize optical figures of merit such as focusing efficiency, achromaticity, Strehl ratio, or image fidelity. We instead optimize a task loss measured after a frozen VLM consumes the sensor output.

Second, end-to-end computational imaging optimizes optical elements for downstream objectives, often jointly with reconstruction, image-processing, or task-specific neural backends~\cite{lin2021end,lin2022end,sitzmann2018endtoend,chang2018hybrid,tseng2021neural,tseng2021differentiable,kienesberger2026end}. Such systems demonstrate the value of task-aware optics, but do not isolate whether a fixed foundation model can be served better by changing only the optical front-end. Recent task-driven lens design is related in motivation but studies a different optical regime, optimizing geometric/refractive lenses for fixed vision models via differentiable ray tracing~\cite{yang2026task}. CODA, by contrast, addresses VLM-aware electromagnetic inverse design of meta-optic front-ends, using Maxwell-based adjoint gradients to update continuous material-density variables.

Third, optical computing and photonic--electronic vision systems use diffractive or metasurface elements as passive computational front-ends for compute offload or edge inference, often with a task-specific digital backend trained around the optical front-end~\cite{lin2018all,colburn2019opticalfrontend,wirthsingh2025apn,peng2026optical}. Recent large-scale metasurface vision systems instantiate a fixed-optics/backend-training direction: the metasurface is reused as an optical feature extractor, while the digital backend is trained on its captured features~\cite{peng2026optical}. CODA inverts this division of labor: the foundation visual encoder is frozen, and only the meta-optic density is optimized, isolating whether the optical front-end alone can adapt to a fixed visual consumer.

\section{Method}
\label{sec:method}
CODA has three components: a frozen-VLM optical objective (Sec.~\ref{sec:problem}), a Maxwell-to-sensor forward model (Sec.~\ref{sec:forward}), and a PSF-interface backward pass that connects frozen-encoder automatic differentiation to an adjoint Maxwell solve (Sec.~\ref{sec:adjoint}).

\subsection{Optimization problem}
\label{sec:problem}

Let $\rho \in [0,1]^N$ denote the design density on a discretized planar meta-optic. We linearly interpolate relative permittivity,

\begin{equation}
 \varepsilon(\rho_i) = \varepsilon_{\min} + \rho_i(\varepsilon_{\max}-\varepsilon_{\min}),
 \label{eq:eps-interp}
\end{equation}
with $\varepsilon_{\min}=1$ and $\varepsilon_{\max}=5.76$. We optimize this continuous grayscale density directly.

Given a frozen VLM $f_\phi$, labeled clean images $\mathcal{D}=\{(I_j,y_j)\}$, and wavelength--angle conditions $\mathcal{C}=\{(\lambda_k,\theta_k)\}_{k=1}^{K}$, let $\mathrm{PSF}_{\mathcal{C}}(\rho)=\{\mathrm{PSF}(\rho;\lambda_k,\theta_k)\}_{k=1}^{K}$ denote the corresponding optical responses. CODA optimizes only $\rho$:

\begin{equation}
 \rho^\star = \arg\min_{\rho\in[0,1]^N}
 \mathbb{E}_{(I,y)\sim\mathcal{D}}
 \mathcal{L}_{\mathrm{CE}}\!\left(
 f_\phi\!\left(\mathcal{A}(I;\mathrm{PSF}_{\mathcal{C}}(\rho))\right), y
 \right),
 \label{eq:problem}
\end{equation}
Here $\mathcal{L}_{\mathrm{CE}}$ is the classification loss, and $\mathcal{A}$ is the differentiable image-formation operator that maps a clean image to the sensor tensor consumed by the frozen model. No learned deconvolution, reconstruction network, or image signal processing module is inserted between the optical front-end and $f_\phi$.

\subsection{Forward model: Maxwell to sensor}
\label{sec:forward}

\begin{figure}[!ht]
\centering
\includegraphics[width=\linewidth]{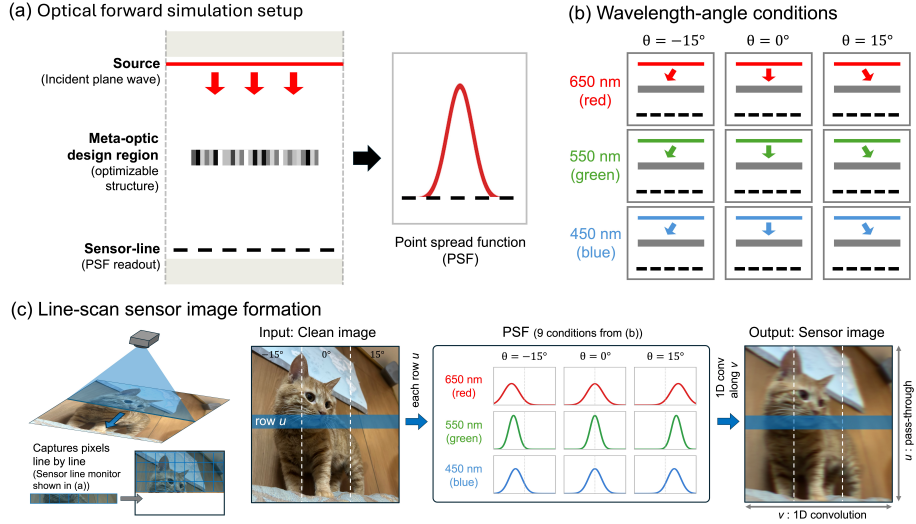}
\caption{
\textbf{Simulation and line-scan image formation.}
(a) A 2D meta-optic simulation uses incident plane-wave illumination, an optimizable meta-optic design region, and a sensor line where the point-spread function (PSF) is measured.
(b) Nine wavelength--angle conditions from three wavelengths and three incidence angles.
(c) The resulting one-dimensional PSFs are applied row-wise to clean images to generate simulated sensor images.
}
\label{fig:simsetup}
\end{figure}

For each wavelength--angle condition $(\lambda_k,\theta_k)$, we compute the two-dimensional TE-polarized electromagnetic response of the meta-optic. We write the target steady-state response as the time-harmonic Maxwell problem for the complex phasor field $E_k$:
\begin{equation}
 \nabla \times \mu_0^{-1}\nabla\times E_k
 - \omega_k^2 \varepsilon_0\varepsilon(\rho)E_k
 = -i\omega_k J_{\lambda_k,\theta_k},
 \label{eq:maxwell}
\end{equation}
where $\omega_k=2\pi c_0/\lambda_k$ and $J_{\lambda_k,\theta_k}$ denotes the incident source associated with wavelength $\lambda_k$ and incidence angle $\theta_k$. Although Eq.~\eqref{eq:maxwell} is written in the frequency domain, the numerical simulation is performed in the time domain. In practice, we use the open-source finite-difference time-domain (FDTD) package Meep~\cite{meep2010} and extract the complex steady-state field component at $\omega_k$; this extracted phasor is denoted by $E_k$ below.

To interface these one-dimensional PSFs with two-dimensional VLM inputs, we use a line-scan approximation, analogous to push-broom acquisition (Fig.~\ref{fig:simsetup}(c))\cite{faraji2019hyperspectral,mouroulis2000design}. 
A 2D cross-section extruded along the invariant axis represents a cylindrical, one-dimensionally focusing meta-optic; Eq.~\eqref{eq:image-formation} implements this approximation by applying the PSF only along the focused horizontal coordinate $v$, treating scan lines at fixed $u$ as optically independent. 

Each wavelength maps to a color channel and each incidence angle maps to a horizontal image zone. For channel $c\in\{R,G,B\}$ and zone $z\in\{\mathrm{left},\mathrm{mid},\mathrm{right}\}$, let $k(c,z)$ index the associated wavelength--angle condition. After resampling each PSF onto the image grid, using the same symbol for brevity, the sensor tensor is
\begin{equation}
 I_{\mathrm{sensor}}[c,u,v]
 = \sum_{v'} I_{\mathrm{clean}}[c,u,v']
 \mathrm{PSF}_{k(c,z(v))}(v-v';\rho),
 \label{eq:image-formation}
\end{equation}
where $u$ and $v$ denote vertical and horizontal image coordinates, respectively, and $z(v)$ maps columns to zones. This approximation keeps the optics--VLM coupling differentiable while avoiding full three-dimensional FDTD over the image field. For computational simplicity, all wavelength--angle conditions are evaluated on the same sensor-line location and represented by PSFs centered on a common readout coordinate. This assumption reduces simulation complexity and isolates the effect of frozen-VLM-aware optical optimization from sensor-geometry considerations.

\subsection{Frozen VLM loss and adjoint-gradient update}
\label{sec:adjoint}

For CLIP and SigLIP, class text embeddings are pre-computed and held fixed. Given a simulated sensor image $I_{\mathrm{sensor}}$, the frozen image encoder produces an embedding $z_{\mathrm{img}}$, and class probabilities are computed from cosine-similarity logits with temperature $\tau$:
\begin{equation}
 p_m = \frac{\exp(z_{\mathrm{img}}^\top z_m^{\mathrm{txt}}/\tau)}
 {\sum_{m'} \exp(z_{\mathrm{img}}^\top z_{m'}^{\mathrm{txt}}/\tau)}.
 \label{eq:softmax}
\end{equation}
The classification loss is $\mathcal{L}_{\mathrm{CE}}=-\log p_y$; below, $\mathcal{L}$ denotes its mini-batch average. The VLM weights and text embeddings are frozen, and only the optical density $\rho$ is updated.

Because the FDTD solver is outside the automatic-differentiation graph, we split the gradient at the PSF interface:
\begin{equation}
\frac{\partial \mathcal{L}}{\partial \rho_i}
= \sum_{k=1}^{K}\sum_x
\frac{\partial \mathcal{L}}{\partial \mathrm{PSF}_k(x)}
\frac{\partial \mathrm{PSF}_k(x)}{\partial \rho_i}.
\label{eq:chain-split}
\end{equation}
The first factor is obtained by reverse-mode automatic differentiation through the frozen VLM and the differentiable image-formation operator. The second is computed with an adjoint Maxwell solve~\cite{Miller:EECS-2012-115,christiansen2021inverse}. For $\mathrm{PSF}_k(x)=|E_k(x,y_f)|^2$, the adjoint source at the sensor-line monitor is
\begin{equation}
 s_k(x)=\frac{\partial \mathcal{L}}{\partial \mathrm{PSF}_k(x)}\,2\overline{E_k(x,y_f)},
 \label{eq:adjoint-source}
\end{equation}
where $2\overline{E_k}$ is the local vector--Jacobian product of the intensity operation under our complex-field convention.
Solving the adjoint Maxwell problem gives an adjoint field $E_k^{\mathrm{adj}}$, yielding

\begin{equation}
 \boxed{
 \frac{\partial \mathcal{L}}{\partial \rho_i}
 = \mathrm{Re}\sum_{k=1}^{K}
 \omega_k^2\varepsilon_0\Delta\varepsilon
 \int_{\Omega_i} E_k(\mathbf{r};\rho)\cdot
 E_k^{\mathrm{adj}}(\mathbf{r};\rho)\,d\mathbf{r}
 },
 \label{eq:adjoint-main}
\end{equation}
where $\Delta\varepsilon=\varepsilon_{\max}-\varepsilon_{\min}$ and $\Omega_i$ is the region of the $i$-th design pixel.

Each update therefore requires $K$ forward and $K$ adjoint FDTD simulations, independent of the number of design pixels. With $K=9$ wavelength--angle conditions, 18 FDTD calls produce gradients for all $N=13{,}400$ density variables; a finite-difference estimate would require $K(N+1)\approx1.2\times10^5$ forward simulations per update.

\begin{figure}[!htb]
\centering
\includegraphics[width=\linewidth]{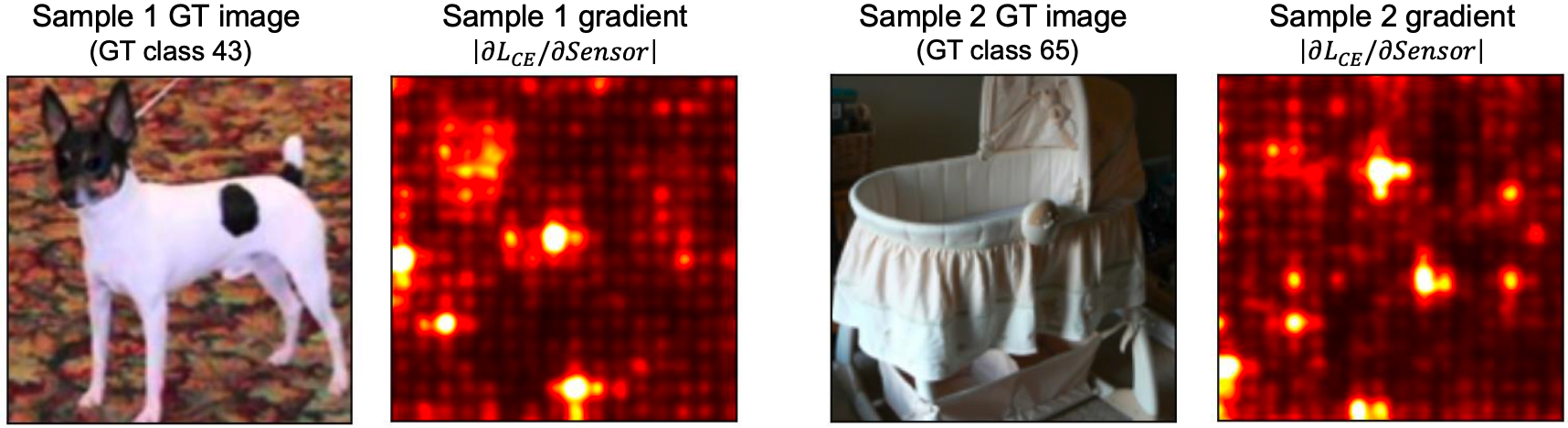}
\caption{
\textbf{Sensor-space gradients induced by the frozen visual classifier.}
For two ImageNet-100 validation images, the heatmaps show the magnitude of the cross-entropy gradient with respect to the simulated sensor image,
$\left|\partial \mathcal{L}_{\mathrm{CE}} / \partial I_{\mathrm{sensor}}\right|$.
These structured gradients are back-propagated through the line-scan image-formation model to produce the point-spread-function gradients used as adjoint sources for the optical update.
}
\label{fig:gradient_vlm}
\end{figure}

\begin{algorithm}[!ht]
\caption{CODA meta-optic optimization with a frozen visual classifier.}
\label{alg:train}
\footnotesize
\begin{algorithmic}[1]
\Require frozen classifier $f_\phi(\cdot;\mathcal{T})$ with fixed class text embeddings $\mathcal{T}$; labeled dataset $\mathcal{D}$; differentiable image-formation operator $\mathcal{A}$, including fixed PSF resampling to the image grid; wavelength--angle conditions $\mathcal{C}=\{(\lambda_k,\theta_k)\}_{k=1}^{K}$; number of iterations $N_{\mathrm{iter}}$; learning rate $\eta$; clip norm $g_{\max}$
\Ensure optimized density $\rho$ (meta-optic front-end design)
\State Initialize density $\rho\in[0,1]^N$ randomly (VLM-cold) or from a Focus-opt checkpoint (VLM-warm)
\State Initialize Adam optimizer state for $\rho$; keep $\phi$ and $\mathcal{T}$ fixed
\For{$n=1,\ldots,N_{\mathrm{iter}}$}
 \State Sample mini-batch $\mathcal{B}=\{(I_j,y_j)\}\subset\mathcal{D}$
 \For{$k=1,\ldots,K$}
  \State $E_k\leftarrow \mathrm{FDTD}(\rho,\lambda_k,\theta_k)$
  \State $P_k\equiv\mathrm{PSF}_k\leftarrow |E_k(\cdot,y_f;\rho)|^2$
 \EndFor
 \State $I_{\mathrm{sensor},j}\leftarrow
 \mathcal{A}\bigl(I_j;\{P_k\}_{k=1}^{K}\bigr)$ for all $(I_j,y_j)\in\mathcal{B}$
 \State $\mathcal{L}\leftarrow |\mathcal{B}|^{-1}
 \sum_{(I_j,y_j)\in\mathcal{B}}
 \mathcal{L}_{\mathrm{CE}}\!\left(
 f_\phi(I_{\mathrm{sensor},j};\mathcal{T}),y_j
 \right)$
 \State Compute $\{\alpha_k\}_{k=1}^{K}$, where
 $\alpha_k\equiv\partial\mathcal{L}/\partial P_k$, by reverse-mode automatic differentiation through $f_\phi$ and $\mathcal{A}$, treating $\{P_k\}_{k=1}^{K}$ as PSF-interface variables
 \For{$k=1,\ldots,K$}
  \State $s_k(x)\leftarrow 2\,\alpha_k(x)\,
  \overline{E_k(x,y_f;\rho)}$
  \State $E_k^{\mathrm{adj}}\leftarrow
  \mathrm{FDTD}_{\mathrm{adj}}(\rho,\lambda_k,\theta_k,s_k)$
 \EndFor
 \State Assemble $g\leftarrow\nabla_{\rho}\mathcal{L}$ from
 $\{E_k,E_k^{\mathrm{adj}}\}_{k=1}^{K}$ using Eq.~\eqref{eq:adjoint-main}
 \State $g\leftarrow\operatorname{clipnorm}(g,g_{\max})$
 \State $\rho\leftarrow
 \mathrm{clip}_{[0,1]}\!\left(\mathrm{AdamStep}(\rho,g,\eta)\right)$
\EndFor
\State \Return $\rho$
\end{algorithmic}
\end{algorithm}

\subsection{Designs compared}
\label{sec:designs}

All optical designs are evaluated under the same $5\,\mu\mathrm{m}\times600\,\mathrm{nm}$ design envelope, FDTD grid, wavelength--angle conditions, and downstream image-formation code. The learned designs use the same density parameterization and optimizer family. We compare four primary designs: Fresnel, an analytical reference; Focus-opt, a model-agnostic focal-concentration baseline optimized for 200 iterations; VLM-cold, CODA from random initialization for 100 iterations; VLM-warm, 100 CODA iterations initialized from the same-seed Focus-opt iteration-100 checkpoint.

The Focus-opt baseline uses no labels, text prompts, or encoder gradients. It optimizes the same density variables with a focal-concentration objective,
\begin{equation}
 \mathcal{L}_{\mathrm{focus}}(\rho)
 = -\frac{1}{K}\sum_{k=1}^{K}
 \frac{\sum_{x\in W_k}\mathrm{PSF}_k(x;\rho)}
 {\sum_x \mathrm{PSF}_k(x;\rho)}
 \label{eq:focus-baseline}
\end{equation}
where $W_k$ is the target focal window for the $k$-th wavelength--angle condition. This baseline represents a model-agnostic optical prior: more energy concentrated near the target focus is expected to yield a better sensor measurement. Its downstream accuracies are evaluated only after optical optimization by passing the resulting sensor tensors through frozen encoders.

VLM-cold and VLM-warm differ only in initialization; both use the CODA gradient path of Eq.~\eqref{eq:chain-split}. VLM-warm gives a budget-matched comparison to 200-iteration Focus-opt: from the same-seed Focus-opt iteration-100 checkpoint, we switch the objective from focal concentration to frozen-VLM cross-entropy for another 100 iterations (Fig.~\ref{fig:trajectory}). Focus-opt, VLM-cold, and VLM-warm each use three seeds. All downstream encoders remain frozen; for DINOv2, which has no text branch, we fit a clean-image linear probe once and keep it fixed across optical designs.

\section{Results}
\label{sec:results}

\subsection{Main ImageNet-100 result}
\label{sec:main-result}

We optimize the structure of a meta-optic front-end on the ImageNet-100~\cite{tian2020crd} training split using CLIP ViT-L/14~\cite{radford2021learning} as the frozen optimization-time consumer, and report all main accuracies on the held-out ImageNet-100 validation split (5{,}000 images). Clean images, evaluated without a simulated optical front-end, reach 88.26\% zero-shot accuracy and serve as the no-optic reference for this pipeline. This clean-image result is not the target to beat; it marks the advantage of the visual domain on which the frozen model is normally trained and evaluated. The relevant comparison is therefore among constrained optical front-ends with the same simulated design envelope and image-formation model. Table~\ref{tab:main} summarizes the designs.

As expected, none of the simulated optical designs reaches the clean-image reference. Within the constrained setting, the analytical Fresnel baseline is severely degraded, showing that a closed-form lens is inadequate in this small-aperture, multi-wavelength configuration. Focus-opt improves accuracy to $53.75\pm3.57$\%, confirming that the model-agnostic focusing baseline is already strong. Direct VLM optimization from random initialization does not improve on Focus-opt: VLM-cold reaches $47.87\pm1.76$\% suggesting that cold-start VLM gradients alone are insufficient to find a useful optical basin. In contrast, VLM-warm reaches $65.41\pm3.99$\%, improving over Focus-opt by $+11.66$ percentage points. The worst VLM-warm seed (61.44\%) exceeds the best Focus-opt seed (57.84\%), so the warm-start advantage is seed-consistent rather than a single favorable run.

\begin{table}[!t]
\centering
\footnotesize
\setlength{\tabcolsep}{5pt}
\renewcommand{\arraystretch}{1.12}
\caption{
ImageNet-100 validation accuracy with frozen CLIP ViT-L/14.
All optical designs use the same simulated design domain and image-formation model; optimized designs report mean $\pm$ standard deviation over three seeds.
Concentration is the mean local sensor-line energy fraction within a 0.5\,$\mu$m target window, averaged over the nine wavelength--angle conditions.
The symbol pp denotes percentage points.
}
\label{tab:main}
\begin{tabular*}{\linewidth}{@{\extracolsep{\fill}}lcc@{}}
\toprule
\textbf{Input / design} & \textbf{Concentration $\uparrow$} & \textbf{Accuracy (\%) $\uparrow$} \\
\midrule
Clean images & -- & 88.26 \\
Fresnel zone plate & 0.114 & 8.10 \\
Focus-opt & $0.432\pm0.021$ & $53.75\pm3.57$ \\
VLM-cold & $0.216 \pm 0.012$ & $47.87 \pm 1.76$ \\
\textbf{VLM-warm (ours)} & $0.425\pm0.013$ & $\mathbf{65.41\pm3.99}$ \\
\midrule
\multicolumn{2}{@{}l}{\textbf{$\Delta$ VLM-warm vs. Focus-opt}} & \textbf{$+11.66$} pp \\
\bottomrule
\end{tabular*}
\end{table}

\begin{figure}[!ht]
\centering
\includegraphics[width=\linewidth]{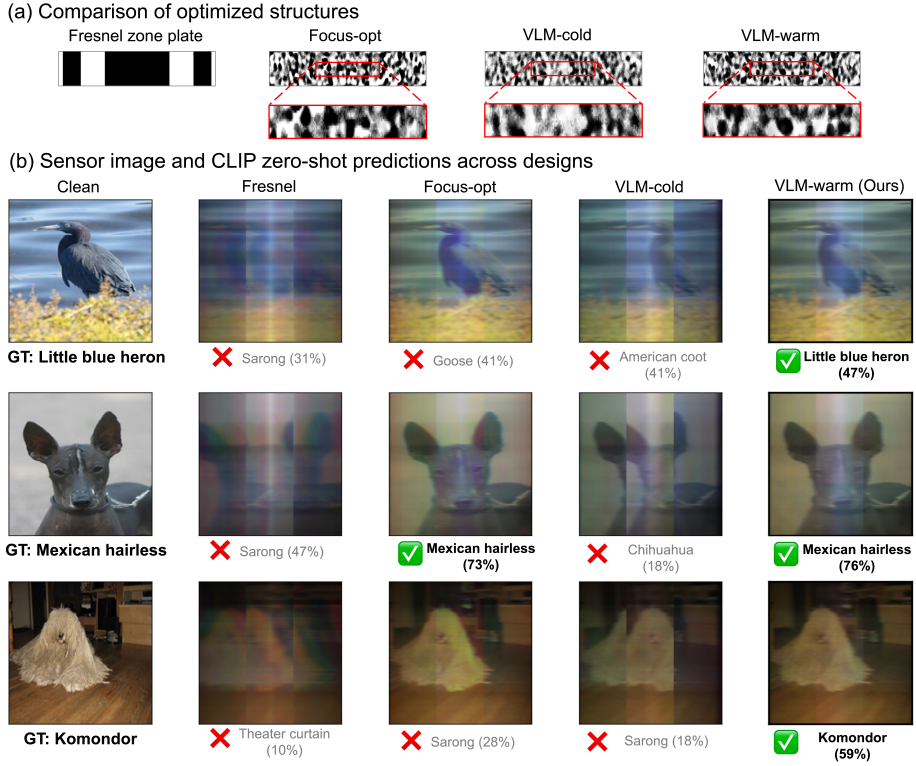}
\caption{
(a) Representative optical designs and fields for the Table~\ref{tab:main} comparison.
(b) Qualitative sensor images and CLIP ViT-L/14 zero-shot predictions on ImageNet-100 validation examples.
Columns compare the clean image with sensor images from the Fresnel zone plate, Focus-opt, VLM-cold, and VLM-warm designs.
Labels show the top-1 prediction and confidence; GT denotes the ground-truth class.
The examples are illustrative, and quantitative claims use the full validation set.
}
\label{fig:sensor_predictions}
\end{figure}

\begin{figure}[!ht]
\centering
\includegraphics[width=\linewidth]{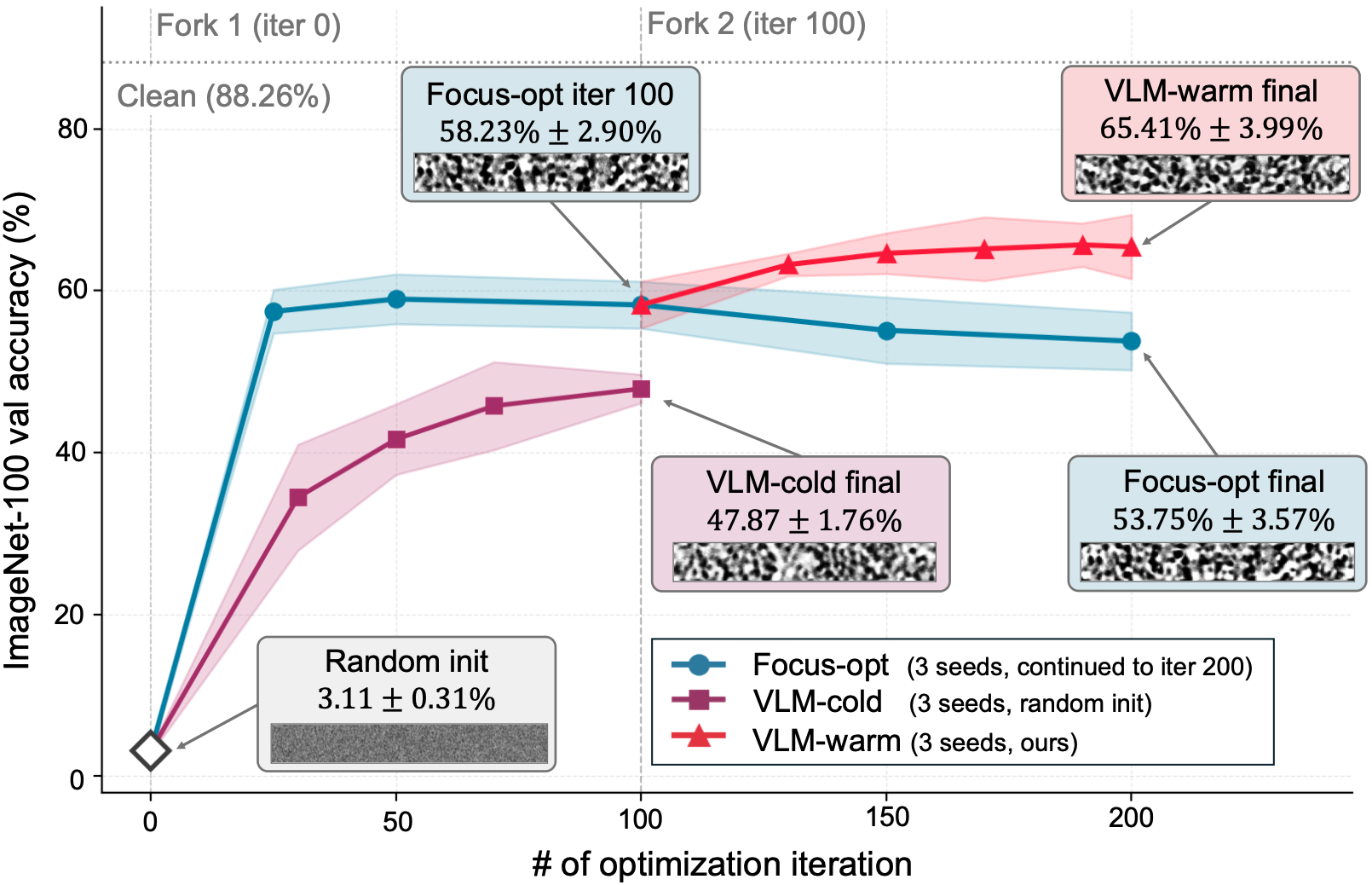}
\caption{
\textbf{Optimization trajectories on ImageNet-100 with frozen CLIP ViT-L/14.}
At iteration~100, VLM-warm forks from the Focus-opt trajectory and switches to the frozen-CLIP classification objective, while Focus-opt continues focal-concentration optimization with the same budget.
VLM-cold uses the frozen-CLIP objective from random initialization.
Shaded bands show $\pm1$ standard deviation over three seeds.
}
\label{fig:trajectory}
\end{figure}

The comparison is also robust to checkpoint selection. The matched-budget comparison uses Focus-opt at iteration~200 because VLM-warm also uses 200 total iterations. However, the best Focus-opt validation checkpoint occurs earlier, at iteration~50, with $58.95\pm3.07$\% accuracy. VLM-warm still exceeds this best Focus-opt checkpoint by $+6.46$ percentage points.

The $+17.54$ percentage-point gap between VLM-warm and VLM-cold indicates that initialization is central to CODA optimization. The two variants share the same optical model, image-formation pipeline, frozen encoder, loss function, and optimizer family; they differ only in the initial density $\rho$. VLM-cold also lands in a basin with substantially worse local focusing: its local peak positions span $-0.20$ to $+0.71\,\mu$m across the nine wavelength--angle conditions, an order of magnitude wider than either Focus-opt or VLM-warm in their corresponding local sensor-line coordinates.

This pattern suggests a difficult VLM cross-entropy landscape over the meta-optic density space. From random initialization, the VLM objective can drive the design toward an
asymmetric local optimum with poor focusing and lower downstream accuracy. Focal-concentration optimization instead moves $\rho$ into a region where the wavelength--angle conditions are jointly well focused, acting as a scaffold for subsequent VLM optimization. Figure~\ref{fig:trajectory} shows this under the same compute budget: from the same iteration-100 density, continuing Focus-opt reaches $53.75$\% by iteration~200, whereas switching to the frozen-VLM objective reaches $65.41$\% over the same 100 additional iterations.

\subsection{Local PSF metrics do not predict recognition gain}
\label{sec:metric-gap}

The VLM-warm gain is not predicted by the local PSF metrics that motivate the Focus-opt baseline. Concentration, full width at half maximum (FWHM), local peak position, and peak intensity all tie or favor Focus-opt, yet VLM-warm improves downstream zero-shot accuracy by $+11.66$ percentage points. Thus, in this constrained optical regime, the metrics that would select Focus-opt under a conventional focusing objective do not select the more accurate front-end for the frozen VLM.

\begin{table}[!ht]
\centering
\footnotesize
\setlength{\tabcolsep}{6pt}
\renewcommand{\arraystretch}{1.15}

\caption{
Local PSF metrics versus downstream accuracy. Compact-focus metrics tie or favor Focus-opt, whereas CLIP zero-shot accuracy favors VLM-warm.
Arrows indicate the conventional preferred direction; $\Delta$ denotes VLM-warm $-$ Focus-opt.
}
\label{tab:metric-paradox}
\begin{tabular*}{\linewidth}{@{\extracolsep{\fill}}lccc@{}}
\toprule
\textbf{Metric} & \textbf{Focus-opt} & \textbf{VLM-warm} & \textbf{$\Delta$} \\
\midrule
Concentration $\uparrow$ & $0.432 \pm 0.021$ & $0.425 \pm 0.013$ & $-0.007$ \\
Median FWHM ($\mu$m) $\downarrow$ & $0.16$ & $0.16$ & $0$ \\
Max local $|x_{\mathrm{peak}}|$ ($\mu$m) $\downarrow$ & $0.03$ & $0.04$ & $+0.01$ \\
Peak intensity (a.u.) $\uparrow$ & $355 \pm 138$ & $229 \pm 101$ & $-126$ \\
\midrule
2nd moment ($\mu$m$^2$) & $1.23 \pm 0.39$ & $1.47 \pm 0.37$ & $+0.24$ \\
\midrule
\textbf{Zero-shot accuracy (\%) $\uparrow$} & $\mathbf{53.75 \pm 3.57}$ & $\mathbf{65.41 \pm 3.99}$ & $\mathbf{+11.66}$ \\
\bottomrule
\end{tabular*}
\end{table}

The main optical difference is PSF spatial extent. VLM-warm has a $20\%$ larger intensity-weighted second moment than Focus-opt, trading lower peak intensity for a broader base. Broader PSFs alone are not sufficient: Fresnel and VLM-cold are also broad or unstable yet perform poorly. We therefore interpret the larger second moment as a correlate of CODA optimization, not as a standalone optical-design rule.

A second line of evidence comes from the frozen encoder embedding space. A 5-fold cross-validated linear probe fit directly on CLIP image embeddings from the same 5{,}000 validation images reaches $73.49\pm3.65\%$ for VLM-warm versus $63.45\pm2.57\%$ for Focus-opt; silhouette and inter-/intra-class distance measures move consistently. Because the probe uses only image embeddings, not CLIP text prompts, the gain is not merely prompt alignment.

\subsection{Transfer without optical re-optimization}
\label{sec:transfer}

We next reuse the same ImageNet-100/CLIP optical designs without optical re-optimization. CLIP and SigLIP are evaluated in the standard zero-shot manner using dataset-specific text prompts. DINOv2 has no text encoder, so for each dataset we fit one linear probe on clean-image DINOv2 training features and then keep that probe fixed across optical designs. Table~\ref{tab:cross-dataset} reports the transfer matrix.

\begin{table}[!ht]
\centering
\footnotesize
\setlength{\tabcolsep}{4pt}
\renewcommand{\arraystretch}{1.12}
\caption{
Transfer without optical re-optimization across datasets and frozen encoders.
The optical designs from the ImageNet-100/CLIP setting are reused unchanged for all entries.
CLIP and SigLIP use zero-shot prompts; DINOv2 uses a frozen encoder with a linear probe fit once on clean-image training features.
Entries report top-1 accuracy; learned optics show mean $\pm$ standard deviation over three seeds, while the Fresnel zone plate is analytical.
$\Delta$ rows show VLM-warm minus Focus-opt in percentage points.
}
\label{tab:cross-dataset}
\begin{tabular*}{\linewidth}{@{\extracolsep{\fill}}llccc@{}}
\toprule
\textbf{Dataset} & \textbf{Design} & \textbf{CLIP (\%)} & \textbf{SigLIP (\%)} & \textbf{DINOv2 (\%)} \\
\midrule
\multirow{5}{*}{ImageNet-100}
 & Fresnel & 8.10 & 1.98 & 5.68 \\
 & VLM-cold & $47.87\pm1.76$ & $32.65\pm3.49$ & $72.52\pm3.77$ \\
 & Focus-opt & $53.75\pm3.57$ & $38.10\pm3.73$ & $77.37\pm2.01$ \\
 & \textbf{VLM-warm} & $\mathbf{65.41\pm3.99}$ & $\mathbf{52.07\pm5.42}$ & $\mathbf{84.98\pm1.04}$ \\
 & $\Delta$ (pp) & $\mathbf{+11.66}$ & $\mathbf{+13.97}$ & $\mathbf{+7.61}$ \\
\midrule
\multirow{5}{*}{CIFAR-100}
 & Fresnel & 8.07 & 5.67 & 4.85 \\
 & VLM-cold & $31.19\pm1.80$ & $22.92\pm2.01$ & $49.91\pm1.32$ \\
 & Focus-opt & $34.45\pm4.36$ & $26.33\pm3.85$ & $52.44\pm4.08$ \\
 & \textbf{VLM-warm} & $\mathbf{51.01\pm5.61}$ & $\mathbf{41.03\pm5.21}$ & $\mathbf{73.24\pm4.73}$ \\
 & $\Delta$ (pp) & $\mathbf{+16.56}$ & $\mathbf{+14.70}$ & $\mathbf{+20.80}$ \\
\midrule
\multirow{5}{*}{Food-101}
 & Fresnel & 2.16 & 2.26 & 1.52 \\
 & VLM-cold & $41.21\pm1.75$ & $24.96\pm1.53$ & $47.09\pm0.85$ \\
 & Focus-opt & $47.35\pm6.50$ & $27.42\pm6.61$ & $49.46\pm8.18$ \\
 & \textbf{VLM-warm} & $\mathbf{65.31\pm7.55}$ & $\mathbf{45.93\pm9.10}$ & $\mathbf{70.26\pm5.14}$ \\
 & $\Delta$ (pp) & $\mathbf{+17.96}$ & $\mathbf{+18.51}$ & $\mathbf{+20.80}$ \\
\bottomrule
\end{tabular*}
\end{table}

VLM-warm outperforms Focus-opt in all nine dataset--encoder pairs, with margins from $+7.61$ to $+20.80$ percentage points. We treat this all-cell win pattern as a consistency check rather than nine independent hypothesis tests, because entries share optical seeds, simulation settings, and related evaluation pipelines. The cold-start ablation reverses the pattern: Focus-opt exceeds VLM-cold in all nine cells by $+2.37$ to $+6.14$ percentage points, confirming that the focusing scaffold matters beyond the ImageNet-100/CLIP optimization cell.

\subsection{Limitations}
\label{sec:limitations}

Our claims are made in a controlled simulation regime. We use two-dimensional FDTD and line-scan image formation to keep frozen-VLM backpropagation tractable at batch size 16 (peak 42\,GB GPU memory). For computational simplicity, all wavelength--angle conditions are evaluated on the same sensor-line location and represented by PSFs centered on a common readout coordinate. Thus, we do not model field-dependent sensor geometry, full two-dimensional wide-FOV image formation, or a complete camera package. The three wavelengths, 450, 550, and 650\,nm, are mapped to RGB channels rather than used to model broadband color, so achromatic or broadband meta-optic design is outside our scope.

Our evaluation protocol also fixes several choices. DINOv2 is evaluated with a clean-image linear probe because it has no text branch, and the probe is fit once and kept fixed across optical designs. Finally, CODA is not meant to outperform clean-image inputs, improve every optical design problem, or replace focusing objectives when faithful image formation is feasible. Our narrower claim is that, for a constrained simulated meta-optic front-end feeding a frozen visual model, VLM-loss-driven adjoint optimization can improve recognition relative to a focal-concentration objective.

\section{Conclusion}
\label{sec:conclusion}

We presented CODA, a meta-optic front-end co-design framework that connects a frozen foundation VLM loss to adjoint-gradient updates of the optical density. The clean-image reference remains the strongest input in our experiments, as expected for a VLM trained on conventionally formed natural images. CODA should therefore be read as a constrained meta-optic front-end result rather than a claim against clean imaging. In our controlled two-dimensional line-scan setting, CODA improves ImageNet-100 CLIP accuracy by $+11.66$ percentage points over Focus-opt under the same simulated design envelope, and the same optimized optics outperform Focus-opt on all nine evaluated encoder--dataset combinations without optical re-optimization. Focal-plane metrics tie or favor Focus-opt, while downstream accuracy and CLIP embedding separability favor VLM-warm. We interpret the warm-start result as evidence that focal-concentration optimization can be a useful scaffold, but that once a constrained optical front-end must feed a frozen VLM, the VLM loss can provide a more effective objective for recognition-oriented optical optimization than focal concentration alone.

\section*{Ethical responsibilities and AI disclosure}
This work is a simulation-only study using public datasets and pretrained vision models as explicit components of the research methodology. It collects no new data, involves no human subjects, and deploys no physical sensing system. The manuscript, analyses, experimental data, results, references, and scientific claims were created and manually validated by the authors. No generative AI tool was used to generate, alter, or fabricate experimental data, quantitative results, references, or scientific claims. GPT Image was used only to render non-data graphical elements in the author-designed conceptual illustration in Fig.~\ref{fig:fig1}.

\bibliographystyle{splncs04}
\bibliography{main}

\end{document}